\pdfoutput=1

\documentclass[11pt,onecolumn]{article}

\usepackage{cvpr}
\usepackage{amsmath, amssymb, graphicx, xfrac, amsfonts, amsbsy, algorithm, color, subfig, url, multirow, hyperref}
\usepackage[noend]{algpseudocode}
\usepackage[section]{placeins}
\usepackage{float}
\cvprfinalcopy

\def \y {\mathbf{y}}
\def \x {\mathbf{x}}
\def \z {\mathbf{z}}
\def \v {\mathbf{v}}

\def \Th {\pmb{\Theta}}

\def \Rspace { \mathbb{R}}

\def \R {\mathcal{R}}
\def \F {\mathcal{F}}

\begin{document}




\title{Deep Spectral CNN for Laser Induced Breakdown Spectroscopy}


\author{ Juan Castorena, Diane Oyen, Ann Ollila, Carey Legget and Nina Lanza }

\maketitle

\begin{abstract}
{\normalfont
This work proposes a spectral convolutional neural network (CNN) operating on laser induced breakdown spectroscopy (LIBS) signals to learn to (1) disentangle spectral signals from the sources of sensor uncertainty (i.e., pre-process) and (2) get qualitative and quantitative measures of chemical content of a sample given a spectral signal (i.e., calibrate). Once the spectral CNN is trained, it can accomplish either task through a single feed-forward pass, with real-time benefits and without any additional side information requirements including dark current, system response, temperature and detector-to-target range. Our experiments demonstrate that the proposed method outperforms the existing approaches used by the Mars Science Lab for pre-processing and calibration for remote sensing observations from the Mars rover, 'Curiosity'. 
}
\end{abstract}


\section{Introduction}
\label{Sec:introduction} 

Laser induced breakdown spectroscopy (LIBS) continues to be one of the core remote sensing technologies with capabilities well suited to acquire the chemical composition of material. 
On Mars, the ChemCam instrument aboard the \emph{Curiosity} rover has acquired more than $800,000$ LIBS signals for the geochemical analysis of rock and soil samples \cite{Wiens:2012}. One of its most important benefits is its capability to investigate even $<$1mm size samples from up to 7m distances. It is equipped with a 1064nm laser and three spectrometers; one for each ultraviolet (UV), visible (VIO) and near infrared (NIR) bands; which altogether are capable of collecting in real-time sample signatures occurring between 240-905nm with a 340-382nm gap. It is through the analysis of such spectral information that qualitative and quantitative chemical content (QQCC) information of an excited sample can be readily extracted. Its effectiveness in querying materials remotely and minimally invasive has motivated the design of a new and more powerful SuperCam instrument aboard the \emph{Perseverance} rover just launched in recent months \cite{NASA:2020}. 

There are however associated difficulties in qualitatively and quantitatively determining the chemical content of materials from LIBS measurements. These are mainly due to both measurement uncertainties introduced by the sensor and, calibration of the spectral signal into the chemical qualitative and quantitative measures of the sample. Sensor measurement uncertainties include the presence of dark current, white noise, continuum due to Bremsstrahlung and recombination radiation \cite{Cremers:2013}, wavelength shifts caused by temperature changes, instrument response effects and changes caused by sensor-to-target range \cite{Wiens:2012}. The associated issues with calibration are related to the couplings between sensor uncertainties and the mixing effects of combinations of chemical elements contained within the target material. 

To mitigate some of these problems, existing methods have focused on pre-processing a spectral signal to reduce or disentangle sensor uncertainties while subsequently and disjointly estimating the calibration functions to convert the pre-processed spectrum into the qualitative and quantitative chemical content of the sample. For example, the work of \cite{Wiens:2013} on Chemcam proposes a multi-step pre-processing sequential strategy based on background subtraction, wavelet Gaussian denoising, continuum removal, match filtering for wavelength offset adjustment, compensation for instrument response, and sensor-to-target distance correction while also disjointly using partial least squares (PLS) \cite{Rasmus:1996} to learn calibration curves from a library of 69 reference rock standards. Such a method requires a priori information of the ambient settings (e.g., temperature, range) and the system characterizations (e.g., instrument response, dark current at a given time and calibration function estimations) to compensate for these problems. Most of such characterizations involve carefully controlled lab acquisition settings, a diverse library of reference standards and a number of acquisitions such is the case for dark current subtraction which uses averages of collections of dark spectrums at a given time. 

The work of \cite{Clegg:2017} seeks to further improve the calibration task independently assuming access to a pre-processed spectrum as obtained in \cite{Wiens:2013} is available. Such method expands the diversity of the library of standards to $408$ and uses a fusion based approach between independent component analysis (ICA) and sub-model partial least squares (SM-PLS) to further refine the qualitative and quantitative chemical estimation performance. It is worth noting that such ICA and SM-PLS methods were shown each to be effective methods for calibrating pre-processed spectrums in the works of \cite{Forni:2013} and \cite{Anderson:2017}, respectively.

More recent work from \cite{Ewusi-Annan:2020}, referred to us while writing this manuscript focuses on the pre-processing task only and compares two machine learning (ML) based approaches. The methods compared are linear PLS and non-linear classical neural nets (NN) both aimed towards learning pre-processing transformations given a raw spectral signal. Although, the study shows the feasibility of non-linear NN's to learn pre-processings and their capabilities to achieve performance comparable to PLS, the NN architecture employed does not exploit the advantages of newer findings. Specially those coming from deep learning (DL) which could help to further improve LIBS performance and computational efficiency.

In this work, we bridge a gap between LIBS and newer deep learning (DL) findings. Such findings have paved the way towards state of the art methods for many signal processing applications including but not limited to speech, audio, vision and graphs \cite{Deng:2014}, and we aim here towards bringing such benefits to LIBS. 
The proposed method is a deep convolutional neural network (CNN) architecture operating on spectral signals (thus dubbed spectral CNN) that learns to both (1) disentangle spectral signal from the sources of sensor uncertainty (i.e., pre-process) and (2) convert the spectral signal into a qualitative and quantitative measure of chemical content (i.e., calibrate) with a few additional stacked regression layers. With an understanding that access to a 'clean' or pre-processed spectrum as in \cite{Wiens:2013} is of importance to the end user for interpretation; we offer the possibility to learn the transformations of (1) and/or (2) both by composition and in an end-to-end fashion. The findings of the proposed method demonstrates low approximation errors to the pre-processings of \cite{Wiens:2013} and superior performance compared to existing techniques in a chemical composition estimate (i.e., calibration) task while also offering additional computational efficiency benefits that make the method ready for real-time deployments.

In what follows, Section \ref{Sec:approach} describes the problem formulations for both pre-processing and calibration tasks while also describes the spectral CNN architectures we propose to solve such problems. Section \ref{Sec:experiments} describes the experimental work and summarizes the results of the proposed method on the 'Mars' and 'Calib' datasets collected by Chemcam \cite{Washu:2020}. Finally, Section \ref{Sec:conclusion} concludes our findings.


\section{Approach}
\label{Sec:approach} 
The proposed approach is a supervised DL based CNN operating on spectral LIBS signals which we refer to in here as 'spectral CNN'. Such spectral CNN learns the tasks of (1) pre-processing and/or (2) estimate a qualitative and quantitative chemical content (QQCC) measure from LIBS spectral signals. To learn how to pre-process LIBS signals given a raw spectral signal, the learning stage at training uses example pairs of raw spectra as input and cleaned pre-processed spectra (using the standard ChemCam method \cite{Wiens:2013}) as labels. For the second task of calibration, we train the spectral CNN with example pairs of raw LIBS signals and their corresponding chemical constituent representation vectors as labels. Once the spectral CNN's have been trained, they can generate either a pre-processed LIBS signal or estimates of its chemical constituents in a single and efficient feed-forward network pass.

In this Section we first formulate the problems characterizing each pre-processing and QQCC tasks. Subsequently, we propose and describe the spectral CNN architectures and functions that dictate the objectives that need to be satisfied at the learning stage.
\begin{figure*}[tbh]
	\centering
	\includegraphics[width=1.00\linewidth]{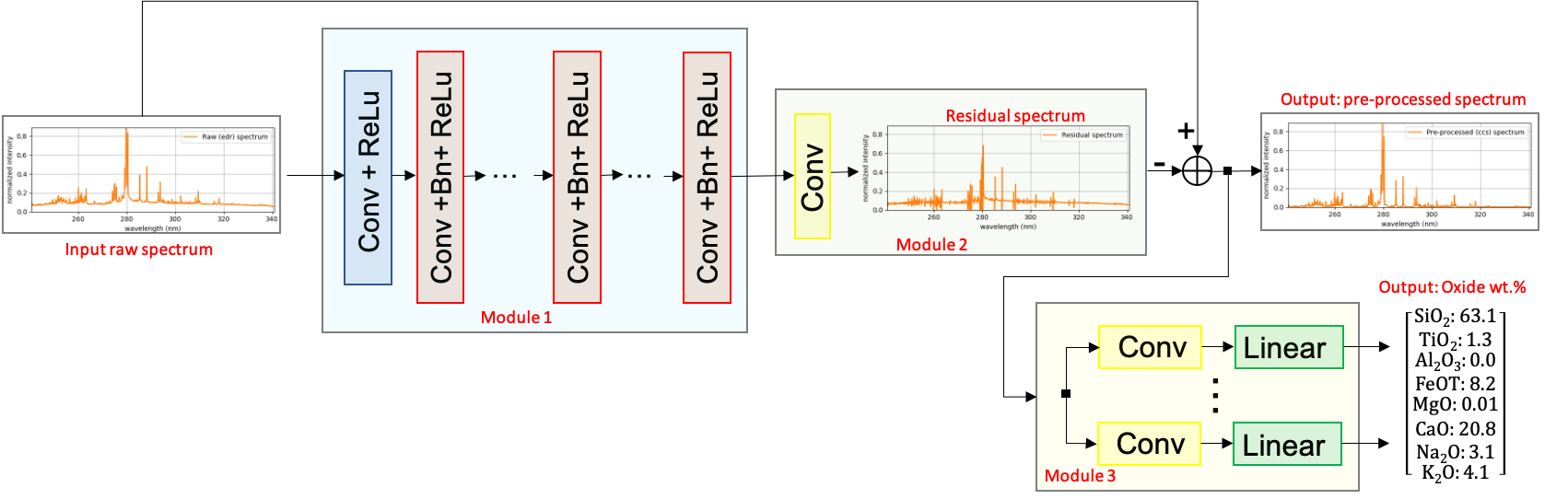}
	\caption{Pre-processing and calibration spectral CNN architectures. The two architectures corresponding to the pre-processing and calibration tasks are constructed by composition of Module-1,2 and Module-1-3, respectively.}
	\label{fig:architecture}
\end{figure*}
%

\subsection{Problem formulation}
\subsubsection{ Pre-processing}
\label{Sssec:learning_preprocessings}

The pre-processing task aims at disentangling the sources of sensor uncertainty and measurement effects from the spectral signal itself. Such sources include the plausible effects of dark background currents, white noise, electron continuum, wavelength offsets, instrument response and target-to-detector range corrections \cite{Wiens:2012}. We begin modeling the effects of dark current background, white noise and electron continuum in LIBS signals using the additive noise model
\begin{equation} \label{noise_model}
	\y = \x + \z.
\end{equation}
Here, $\y \in \Rspace^N$ is the the raw measured LIBS signal, $\x \in \Rspace^N$ is the clean signal and $\z \in \Rspace^N$ is the noise including dark current, ‘white’ noise and background continuum. Although not explicit; such noise model is also used in \cite{Wiens:2013} and can be derived through the sequential application of dark current subtraction, white noise denoising and background continuum removal methods therein.
	
The problem of denoising or cleaning in Eq. \eqref{noise_model} is tackled here by learning a function that discriminates the LIBS signal $\x$ from the sensor noise $\z$. In some sense, such formulation assumes the distribution of sensor noise has a structure given the measurements $\y$ that is exploited by learning. We propose to do this in a residual learning framework through a deep spectral convolutional neural network (CNN) that predicts a residual function $\R: \Rspace^N \times \Rspace^{L_1} \rightarrow \Rspace^N$ satisfying:
\begin{equation} \label{preprocessing-model}
	\y = \x + \R(\y, \Th_1).
\end{equation}
where comparing with Eq. \eqref{noise_model} implies $\hat{\z} = \R(\y, \Th_1)$.  Such residual framework (as opposed to directly estimating the clean signal $\x$) gives in a DL context the flexibility to use high performance networks with lower approximation error and faster training rate capabilities \cite{He:2016}. Learning weights $\Th_1 \in \Rspace^{L_1}$ is performed through the optimization of a loss function using a training set of $M$ raw-clean (level1a) signal example pairs $\{ ( \y_m, \x_m )\}_{m=1}^M$. This loss function $f_1:  \Rspace^{L_1} \rightarrow \Rspace$ is the average $\ell_2$-norm error between ground truth and estimation residuals described by:
\begin{equation} \label{preprocessing_error}
f_1(\Th_1) = \frac{1}{M} \sum_{m=1}^{M} \|        
\underbrace{  \y_m - \x_m }_{ \z_m }
- \R(\y_m, \Th_1) \|_{\ell_2}.
\end{equation}
Once the weights $\Th_1$ are learned, the spectral CNN gradually removes the clean signal $\x$ from the input signal $\y$ as it goes through the layers of the network in feed-forward mode.
	
The spectral CNN architecture used is shown in Figure \ref{fig:architecture} and composed of Module-1 and Module-2. Module 1 consists of a convolution and rectifier linear unit (ReLu) layer followed by a sequence of $D$ convolution, batch normalization (Bn) and ReLu interleaved layers while Module-2 is a convolution and subtraction of the residual estimation. A similar architecture than the one used for pre-processing here was proposed in the context of image denoising in \cite{Zhang:2017} and resulted in state of the art performance over sophisticated handcrafted methods such as the BM3D \cite{Dabov:2006} and denoising by soft-thresholding in a wavelet decomposition \cite{Donoho:1995}.

Additional pre-processings of the LIBS signal such as compensation for the modulation induced by the instrument response function (IRF) is also addressed here in a residual based formulation. The model, loss function and architecture is the same as those described in Eq. \eqref{preprocessing-model} and \eqref{preprocessing_error} and Figure \ref{fig:architecture}, respectively. However, the difference lies in the training stage of the spectral CNN where instead of training with raw-clean(level1a) signal pairs we train with $M$ raw-clean (level1b) signal example pairs $\{ ( \y_m, \x_m )\}_{m=1}^M$. Note though, that we do not place attention into modifying the signal axis to be in the appropriate scale of photons per DN (intensity) as in \cite{Wiens:2013}. Reason being that in a multivariate analysis approach and from an information theoretic standpoint scaling a LIBS signal does not bring any new information that can further improve signature analysis. We thus make our framework invariant to scaling and additional normalizations in \cite{Wiens:2013} by normalizing raw and pre-processing signals to be of unit $\ell_2$-norm.
As a final note to this subsection we would like to mention that our framework is flexible to include only the user desired pre-processings. For example, if the user wants to independently collect a number of dark current signals and subtract them separately this can be done by training the network with the appropriate pairs  $\{ ( \y_m, \x_m )\}_{m=1}^M$ omitting dark current subtractions. With regards to wavelength adjustments, we omit this step from our framework as it is related to the association of wavelengths to the independent axis index of the LIBS signal and instead recommend usage of the method proposed in \cite{Wiens:2013}.

\subsubsection{ Chemical qualitative and quantitative estimates.}
\label{Ssec:chemometric_mappings}
%
Further processing aiming at getting the chemical element constituents from the LIBS signal signatures is also treated here. This problem, also referred to as calibration in the LIBS context consists in the estimation of a vector $\v \in \Rspace^C$ representing each entry a chemical compound and the value at each entry a quantitative measure or estimate of the relative to $\%$-oxide percentage. The number $C$ represents the total number of chemical elements to be determined and is user-defined. The problem is generalized thus here as finding an operator $\F: \Rspace^N \rightarrow \Rspace^C$ that takes as input a LIBS signal $\x \in \Rspace^N$ and outputs the vector $\v$ of chemical composition/concentration as given in Eq. \eqref{calibration-model} as:
\begin{equation} \label{calibration-model}
	\v = \F( \x ) 
\end{equation}
Similar to \cite{Clegg:2017}, we assume a linear regression calibration model and let $\F(\x, \Th_2)$ be learned by a shallow neural network with weights $\Th_2 \in \Rspace^{L_2}$. The  neural net employed consists of a set of convolutional and fully connected (FC) layers connected as shown in Module-3 of Figure \ref{fig:architecture} which regress LIBS signatures to chemical content.

Learning the weights $\Th_2$ is done in training with example pairs $\{ (\x_m, \v_m)\}_{m=1}^M$ of pre-processed LIBS signals and corresponding chemical constituents. The optimization minimizes the loss function $f_2: \Rspace^{L_2} \rightarrow \Rspace$ measuring the average $\ell_2$-norm error between a vector representing the ground truth element concentration labels $\v_m \in \Rspace^C$ and the vector of estimates $\hat{\v}_m \in \Rspace^C$ produced by the network. Such loss function can be represented as given by Eq. \eqref{calibration_error}:
\begin{equation} \label{calibration_error}
f_2( \Th_2) = \frac{1}{M} \sum_{m=1}^{M} \| \v_m - \underbrace{   \F(\x_m, \Th_2)  }_{\hat{\v}_m }  \|_{\ell_2}.
\end{equation}
%

The input $\x_m$ in \eqref{calibration_error} is the pre-processed spectrum  obtained in a feed-forward pass through the trained spectral CNN as in Section \ref{Sssec:learning_preprocessings} or any other suitable method such as \cite{Wiens:2013}. We refer to this type of training as training by composition as it is composed of two independent stages: (1) for pre-processing and (2) for calibration both trained independently and under supervision. The overall architecture pipeline is illustrated in Figure \ref{fig:architecture}. One of the benefits of such training scheme is that it offers access to the intermediate pre-processed signal which can be useful and/or necessary for user interpretation. However, this may come with a cost in performance in that it is expected to have a pre-processing quality no better than that of the training pre-processed ground truth signals $\x_m$. 

We also propose a training scheme were the network architecture in Figure \ref{fig:architecture} learns the meaningful LIBS signatures without supervision and in an end-to-end fashion. In such a scheme the network learns a function $\F': \Rspace^N \times \Rspace^{L_3} \rightarrow \Rspace^C $ with weights $\Th_3 \in \Rspace^{L_3}$ that takes a raw LIBS signal $\y \in \Rspace^N]$ as input and estimates its chemical constituents $\v \in \Rspace^C$ as output. The spectral CNN is trained with backpropagation using raw LIBS and chemical constituent example pairs $\{(\y_m, \v_m)\}_{m=1}^M$. The optimization seeks to minimize the average $\ell_2$-norm error between ground truth and estimates as in Eq. \eqref{calibration_error2}
\begin{equation} \label{calibration_error2}
f_3( \Th_3) = \frac{1}{M} \sum_{m=1}^{M} \| \v_m -  \F' (  \y_m, \Th_3 )  \|_{\ell_2}.
\end{equation}
The benefits of such end-to-end training scheme are two-fold: (1) it no longer requires supervision for pre-processing and instead lets the network learn the optimal solution in the sense of \eqref{calibration_error2} and (2) we can achieve lower error as we do not introduce additional noise to the system through the estimation of $\x_m$ as in Eq. \eqref{calibration_error}. In the next section, we validate the proposed approach empirically in both pre-processing and calibration tasks with data obtained by Chemcam in both Mars and earth lab environments.
\section{Experiments} 
\label{Sec:experiments}

The ChemCam datasets available at \cite{Washu:2020} were used as the experimentation benchmark for validation of the proposed approach. The datasets contains corresponding dark, raw, level1a, level1b pre-processed individual spectra obtained from different sols, multiple laser shots and rocks, soil, and calibration standards of varying chemical content. The method used in these datasets to obtain level1a and level1b pre-processed spectral signals follows the methodology described in \cite{Wiens:2013}. The specific datasets we employ are two: (1) shot spectrum measurements from the Mars dataset up to sols 500 and (2) shot spectrum measurements from the reference standard calibration dataset \cite{Clegg:2017}. The first dataset referred to in here as 'Mars' was collected on Mars while the second dataset referred to as the 'Calib' dataset was obtained in a laboratory setting on Earth using 408 calibration standards of known and certified qualitative and quantitative chemical reference composition. In the 'Calib' dataset all wavelengths within the bands [240.811,246.635], [338.457,340.797], [382.13,387.859], [473.184,492.427], [849,905.574] were filtered out consistently throughout all shots as suggested in \cite{Clegg:2017}. It is worth noting that the experimental settings (e.g., focus) and the spectral data itself in \cite{Washu:2020} has been evaluated to be of high quality prior to its upload and online availability. 

The spectral CNN architecture used to pre-process LIBS signals as described in Section \ref{Sssec:learning_preprocessings} consists of a sequence of $D=20$ Conv+Bn+ReLu layers. In each layer, we use 64 convolution filters each of size $3 \times 3$ totalling an effective receptive field  of $41 \times 41$. For learning, we used Adam optimizer with batch sizes of 16 and number of epochs 20. Such selection of parameters resulted in the best trade-off between performance and computational complexity both at training and testing in our experiments. The architecture used to learn qualitative and quantitative chemical content also uses a sequence of 20 Conv+Bn+ReLu layers but with stacked convolution and FC layers as shown in Module-3 of Figure \ref{fig:architecture}. The number of parallel channels in Module-3 were set to 8, these corresponding to the total number of major chemical elements for which a qualitative and quantitative measure is to be estimated. Such number is scalable to any number of elements one wants to estimate chemical composition for.

\subsection{Pre-processing spectral signals.} \label{SSec: learningMappings}
Experiments in this subsection validate the capability of the proposed architecture to learn to pre-process LIBS signals. For this task, we evaluate pre-processing of individual shots at the levels 1a and 1b in \cite{Wiens:2013}'s convention. The 'Mars' dataset on which we validate our approach is partitioned into randomly drawn $\sim75K$ and $\sim12K$ spectrum shots taken during sols 10-400 and used as the training and testing subsets, respectively. The 'Calib' dataset is partitioned into randomly drawn $\sim90K$ and $\sim10K$ shots for the training and testing subsets, respectively.  In terms of computational demands, the architecture we employ for pre-processing requires optimization of $\sim4K$ parameters and $\sim1MB$ of memory space. All single shot spectral signals in the datasets were normalized to be of maximum intensity one, with exceptions when testing where additional normalizations to unit $\ell_2$-norm were used for comparison. Also. note that each shot is of dimension $N \sim 5500 $ (after removal of the unused wavelength band gaps) thus containing each shot signal that many number of spectra.

\subsubsection{Level 1a: learning background subtraction, gaussian denoising, continuum removal and wavelength adjustments.} \label{SSSec:level1a}
\begin{figure} [H]
	\centering 
	\subfloat[Individual shot raw spectrum] { \includegraphics[width=1.00\linewidth]{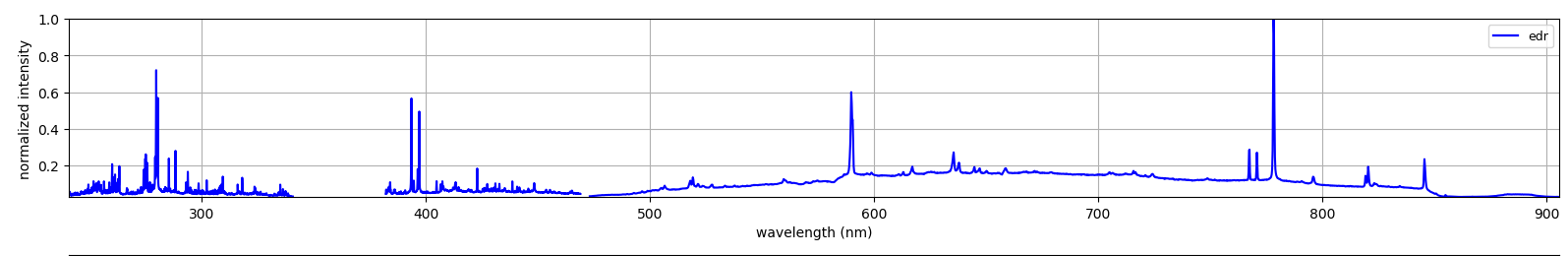} }
	
	\subfloat[Individual shot level 1a pre-processed spectrum comparison.]{ \includegraphics[width=1.00\linewidth]{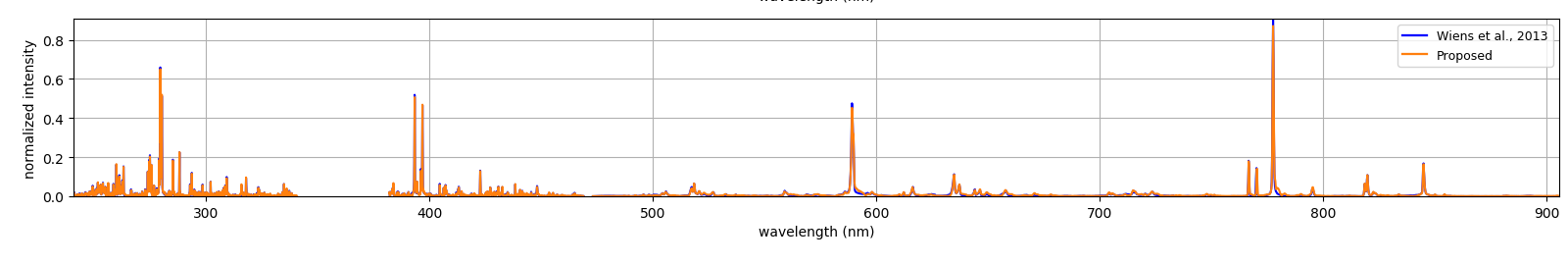} }
	
	\subfloat[Level 1a: Ultraviolet]{ \includegraphics[width=1.00\linewidth]{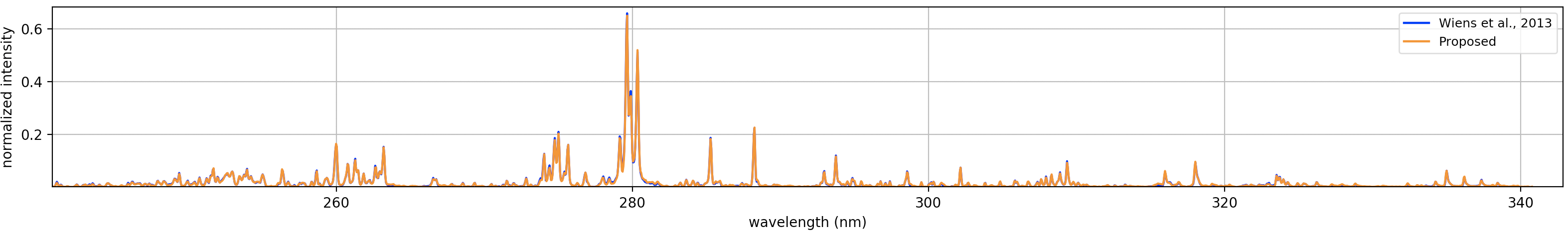} }
	
	\subfloat[Level 1a: Visible]{ \includegraphics[width=1.00\linewidth]{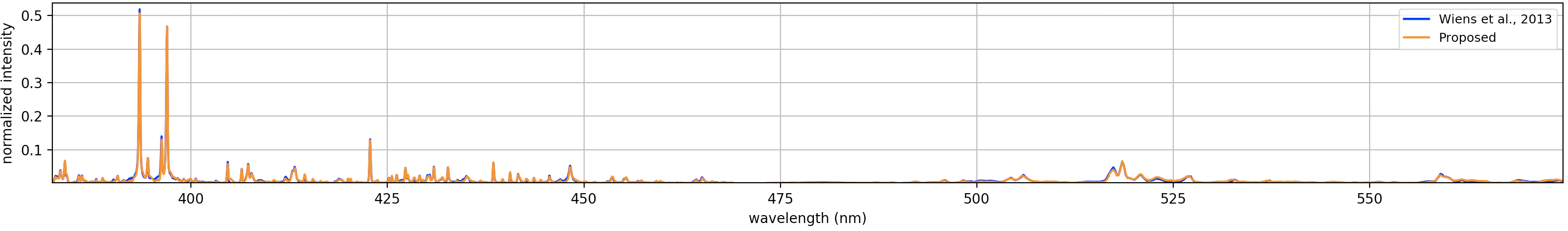} }
	
	\subfloat[Level 1a: Visible]{ \includegraphics[width=1.00\linewidth]{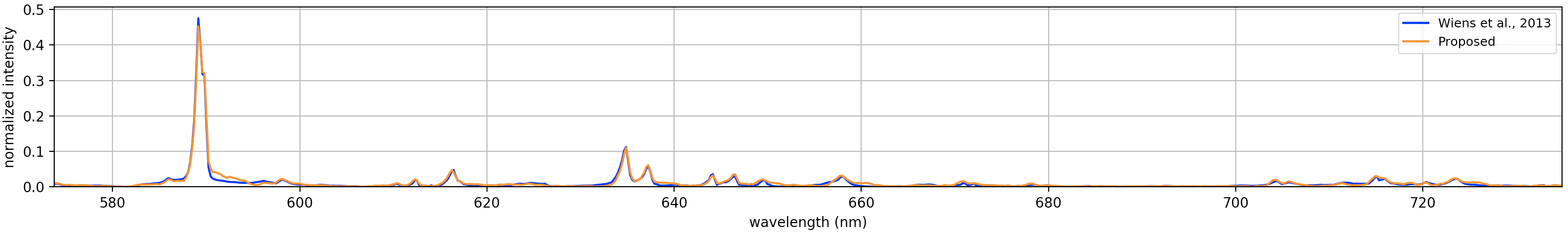} }
	
	\subfloat[Level 1a: Near infrared]{ \includegraphics[width=1.00\linewidth]{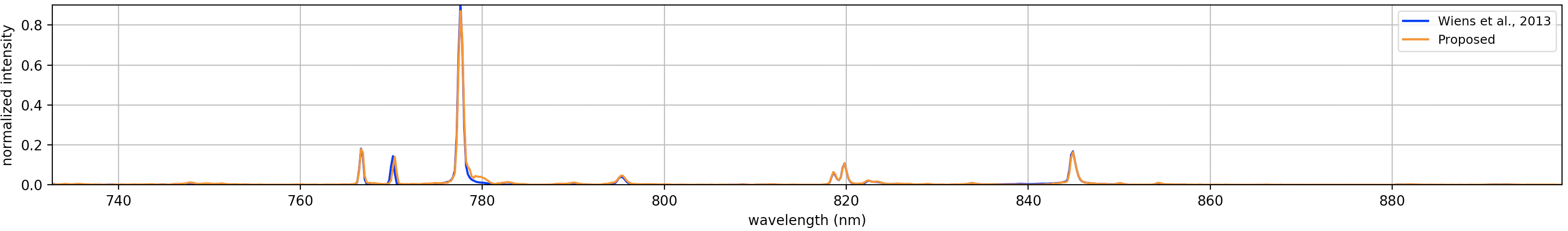} }
	\caption{ Pre-processing example (sols:0293, target:Duluth, dist:2.68m) from 'Mars' dataset: raw to pre-processed (level1a or rdr in \cite{Wiens:2013}). Proposed spectral CNN was capable of learning accurate approximations (RMSE=0.018) to the multi-step method of \cite{Wiens:2013} using as input only a single shot's raw spectrum.}
	
	\label{fig:learning_1a} 
\end{figure}

The training stage aimed to learn raw (edr)-to level1a (rdr) pre-processings as described in Section \ref{Sssec:learning_preprocessings} uses as the training set the raw and level1a corresponding pairs as input and label, respectively. Such level1a pre-processing includes background subtraction, white noise denoising and continuum removal. A representative example of a result and comparison against that of \cite{Wiens:2013} in the test 'Mars' dataset is included in Figure \ref{fig:learning_1a}.
Note that in general, the proposed method was able to learn and predict the pre-processing method of \cite{Wiens:2013} well, without the prior information requirements (e.g., dark current acquisitions and estimates) while also coping well with the differences among all three detector types (i.e., UV, VIS, NIR). In general, most of the peaks obtained by the pre-processing in \cite{Wiens:2013} where well approximated by the proposed deep spectral CNN. For reference, the $\ell_2$-norm error between \cite{Wiens:2013} and the proposed method for the specific case shown in Figure \ref{fig:learning_1a} is  0.018 with a maximum unit $\ell_2$-norm error normalization.

\subsubsection{Level 1b: level 1a +  instrument response, normalization, and distance correction.} \label{SSSec:level1b}
\begin{figure} [H]
	\centering 
	
	\subfloat[Raw (edr) spectrum signal] { \includegraphics[width=1.00\linewidth]{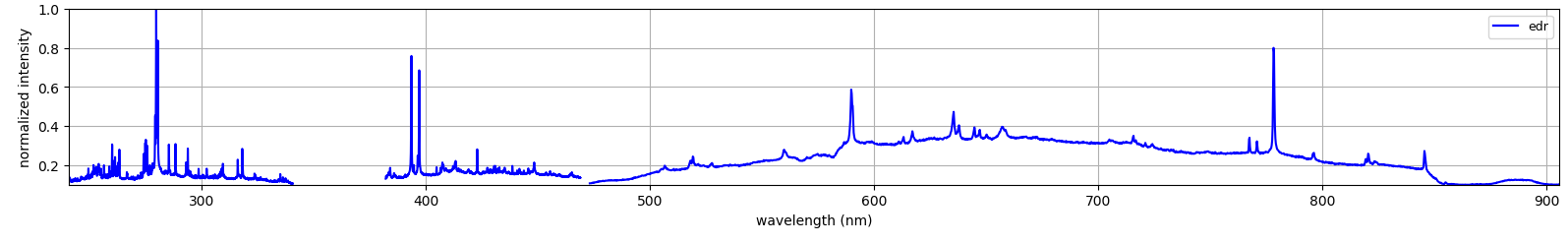}}
	
	\subfloat[Pre-processed (ccs) spectrum signal comparison] { \includegraphics[width=1.00\linewidth]{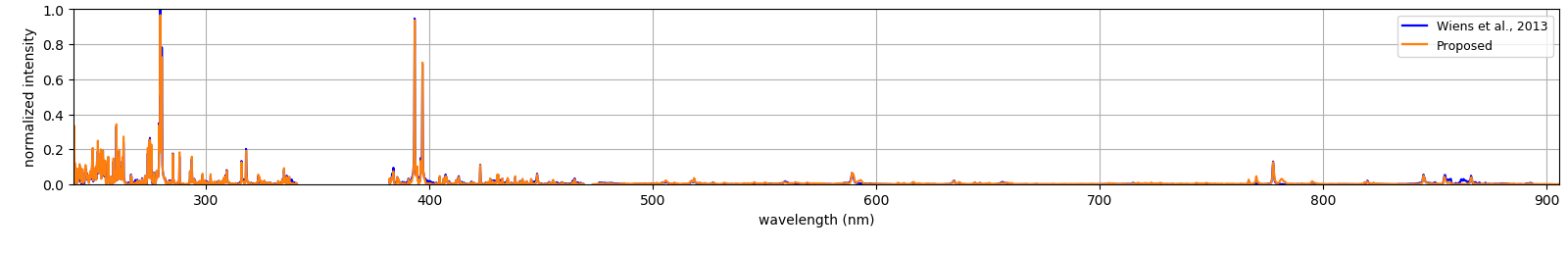}}
	
	\subfloat[UV zoomed pre-processed (ccs) spectrum signal comparison] { \includegraphics[width=1.00\linewidth]{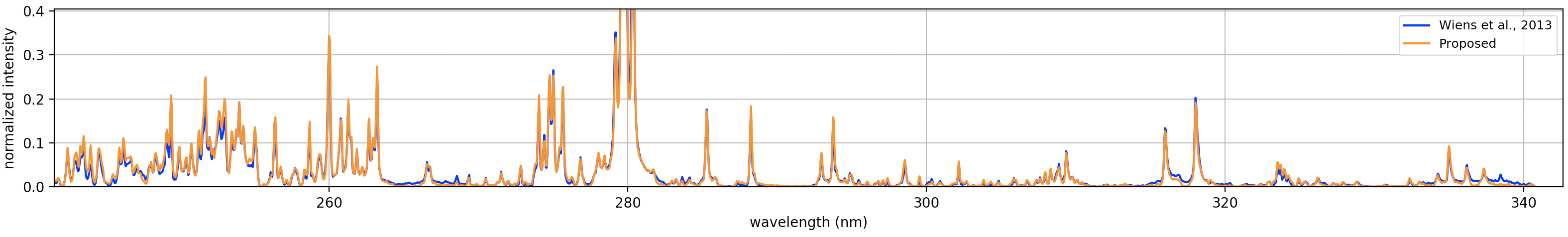}}
	
	\subfloat[VIS zoomed pre-processed (ccs) spectrum signal comparison] { \includegraphics[width=1.00\linewidth]{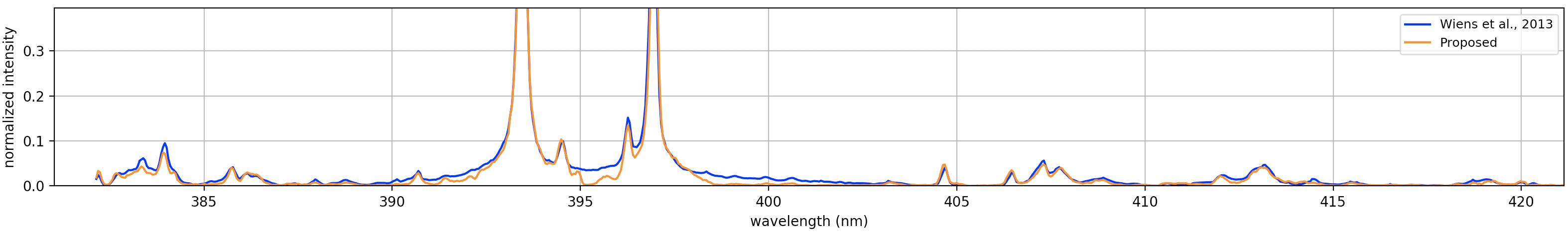}}
	\caption{Pre-processing example (sols:0234, target:McGrath, dist:2.28m) in 'Mars' dataset: raw to pre-processed (level1b or ccs ). Spectral CNN is able to learn how to remove dark current, electron continuum, white noise, and perform adjustments for IRF for all three UV, VIS, NIR detectors, wavelength offset and range. RMSE between \cite{Wiens:2013} and deep CNN is 0.026.}
	
	\label{fig:learning_1b_ex1} 
\end{figure}

\begin{figure} [H]
	\centering 
	
	\subfloat[Raw (edr) spectrum signal ] { \includegraphics[width=1.00\linewidth]{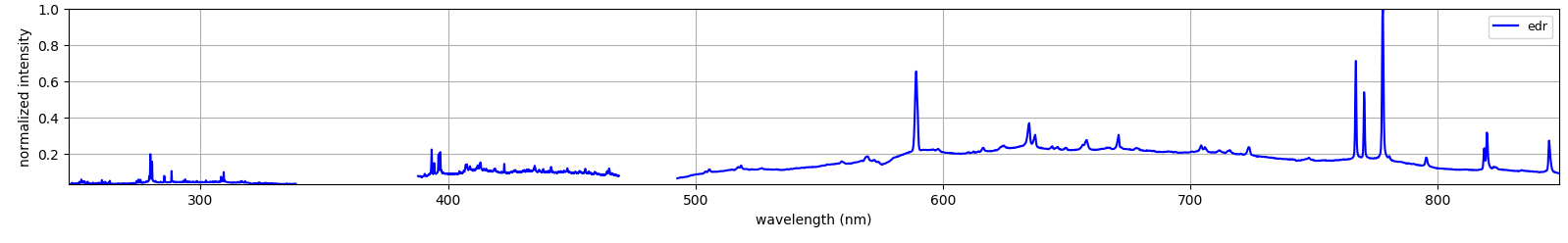}}
	
	\subfloat[Pre-processed (ccs) spectrum signal ] { \includegraphics[width=1.00\linewidth]{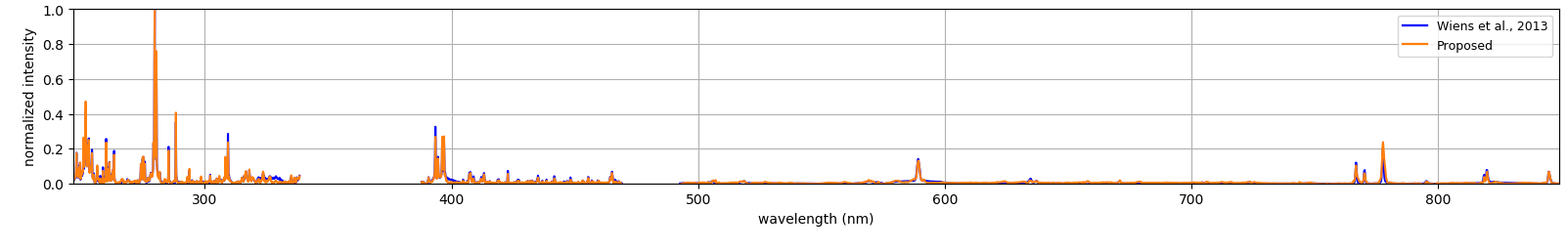}}
	
	\subfloat[UV zoomed pre-processed (ccs) spectrum signal ] { \includegraphics[width=1.00\linewidth]{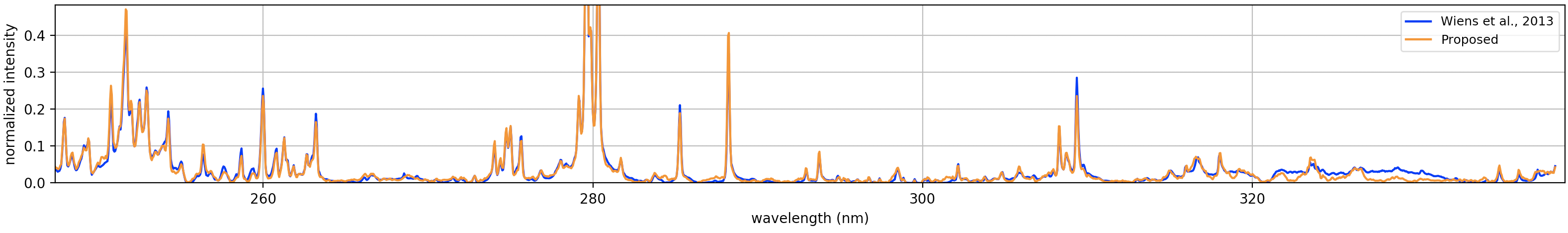}}
	
	\subfloat[VIS zoomed pre-processed (ccs) spectrum signal ] { \includegraphics[width=1.00\linewidth]{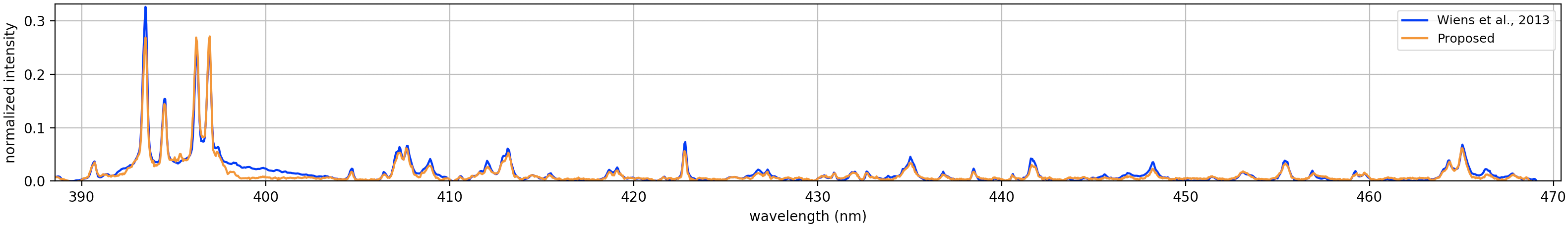}}
	\caption{Pre-processing example (Target:r2100, shot:7, dist:1.6m) in 'Calib' dataset: raw to pre-processed (level1b or ccs). Spectral CNN is able to learn how to remove dark current, electron continuum, white noise, and perform adjustments for IRF for all three UV, VIS, NIR detectors, wavelength offset and range. RMSE between \cite{Wiens:2013} and spectral CNN is 0.068.}
	
	\label{fig:learning_1b_ex2} 
\end{figure}

Learning to pre-process at level1b which in addition to including the steps of level1a accounts for instrument response function (IRF) adjustments and other data normalizations to account for, for example, sensor-to-target distance was carried out as a task by the same approach described in \ref{Sssec:learning_preprocessings}. The process to learn such pre-processing consists on training the network with raw (edr) and level1b (ccs) corresponding pairs. Experiments were conducted on both 'Mars' and 'Calib' datasets, independently. Representative example results from the test 'Mars' and 'Calib' datasets are included in Figure \ref{fig:learning_1b_ex1} and Figure \ref{fig:learning_1b_ex2}, respectively. 
In both cases, the proposed spectral CNN is capable of learning and subtracting good approximations to the background dark spectrums of \cite{Wiens:2013}. This with the additional advantage of removing the dark currents spectrum collection requirements prior to material sampling. In addition, as is illustrated specially in Figure \ref{fig:learning_1b_ex1}.d, the proposed method seems to be better at reducing the overlap between neighboring peaks than \cite{Wiens:2013} a matter indicative of a more effective electron continuum removal. The spectral CNN was also capable of learning and effectively adjusting for the IRF for all three detectors (UV, VIS, NIR) similar to the methodology in \cite{Wiens:2013}. For reference, the $\ell_2$ difference between \cite{Wiens:2013} and the proposed spectral CNN pre-processing is of 0.026 and 0.068 for Figure \ref{fig:learning_1b_ex1} and Figure \ref{fig:learning_1b_ex2}, respectively, each with a max unit $\ell_2$-norm error normalization.

Table \ref{tab:table1} summarizes the root mean squared error (RMSE) expectation overall examples in the test datasets in both 'Mars' and 'Calib' cases and at the two pre-processing levels experimented with, in this paper. Here, RMSE  is measured by \eqref{preprocessing_error} with individual spectra normalized to unit $\ell_2$-norm. 
\begin{table} [H]
	\caption{\label{tab:table1} Performance comparison for pre-processing task as measured by RMSE.}
	\centering
	\resizebox{0.5 \columnwidth}{!}{
	\begin{tabular}{lcccc}
		\hline
		& \multicolumn{2}{c}{Mars} &  \multicolumn{2}{c}{Calib}   \\
		\multirow{2}{*}{Method} & \multirow{2}{*}{Level 1a} &  \multirow{2}{*}{Level 1b}   & \multirow{2}{*}{Level 1a} &  \multirow{2}{*}{Level 1b} \\
		& & & & \\
		\hline
		\\
		Proposed & \multirow{2}{*}{0.053} & \multirow{2}{*}{0.078} & \multirow{2}{*}{N/A} & \multirow{2}{*}{0.079} \\
		Spectral CNN &  &  & & \\
		
		\hline
	\end{tabular}
}
\end{table}
Note that in all cases the proposed method shows relatively small RMSE values meaning the learned mappings are generally good approximations to the multi-step pre-processings done as in \cite{Wiens:2013}. Note that in Table \ref{tab:table1}, Level 1a pre-processing results in the 'Calib' dataset appears as 'N/A' since labels were not available for such a case. In addition to these results, we also evaluate for any degradation in pre-processing performance as a function of detector-to-target distance. Table \ref{tab:table3} includes the RMSE value results of the pre-processing performance as a function of source to target physical distance where distance is measured in meters (m). Note that such test includes only level 1b (ccs) pre-processings on the 'Mars' dataset since the 'Calib' dataset was obtained at a fixed distance throughout all standards.
\begin{table} [ht]
	\caption{\label{tab:table3} Level 1b (ccs) pre-processing RMSE performance versus sensor-to-target distance in meters (m).}
	\centering
	\resizebox{0.5 \columnwidth}{!}{
		\begin{tabular}{lcccccc}
			\hline
			\\
		    'Mars' data &  \multirow{2}{*}{ \pmb{1-2(m)}}   & \multirow{2}{*}{\pmb{2-3(m)}} & \multirow{2}{*}{\pmb{3-4(m)}} & \multirow{2}{*}{\pmb{4-5(m)}} & \multirow{2}{*}{\pmb{5-6(m)}} & \multirow{2}{*}{\pmb{6-7(m)}} \\
            sols:10-500 & & & & & & \\
		    \hline
            \\
            Number of & \multirow{3}{*}{4,055} & \multirow{3}{*}{39,445} & \multirow{3}{*}{24,598} & \multirow{3}{*}{6,884} & \multirow{3}{*}{2,730} & \multirow{3}{*}{682}\\
		    examples & & & & & & \\
		    (train) & & & & & & \\     
		    \\
		    \multirow{2}{*}{RMSE (train)} & \multirow{2}{*}{0.081} & \multirow{2}{*}{0.074} & \multirow{2}{*}{0.074} & \multirow{2}{*}{0.075} & \multirow{2}{*}{0.077} & \multirow{2}{*}{0.075} \\
		    & & & & & & \\
             \\
    		Number of & \multirow{3}{*}{465} & \multirow{3}{*}{4,400} & \multirow{3}{*}{2,692} & \multirow{3}{*}{756} & \multirow{3}{*}{310} & \multirow{3}{*}{88}\\
    	     examples & & & & & & \\
           (test) & & & & & & \\
     		\\
           \multirow{2}{*}{RMSE (test)} & \multirow{2}{*}{0.079} & \multirow{2}{*}{0.075} & \multirow{2}{*}{0.079} & \multirow{2}{*}{0.076} & \multirow{2}{*}{0.089} & \multirow{2}{*}{0.085}\\
		   & & & & & & \\
		   \hline
		\end{tabular}
 	}
\end{table}
%


%
\begin{table*} [t]
	\caption{\label{tab:table2} Performance comparison for the QQCC of the 8 major elements as measured by RMSE in oxide wt.$\%$.}
	\centering
	\resizebox{0.8 \columnwidth}{!}{%
	\begin{tabular}{c|cccccc}
		\hline
		\multirow{3}{*}{Element} & \multirow{2}{*}{pre-Proc.\cite{Wiens:2013} } & \multirow{2}{*}{pre-Proc.\cite{Wiens:2013} } &  \multirow{2}{*}{pre-Proc.\cite{Wiens:2013} } & \multirow{2}{*}{Composition} & \multicolumn{2}{c}{end-to-end} \\
		& & & & & \multicolumn{2}{c}{spectral CNN} \\
		& SM-PLS\cite{Anderson:2017, Clegg:2017} & ICA \cite{Forni:2013, Clegg:2017}& linear NN reg. &  spectral CNN & partition I & partition II\\
		\hline
		& & & & & \\
		SiO$_2$ & 4.33 & 8.31 & 4.897 & 4.716 & 2.859 & 4.193 \\
		TiO$_2$ & 0.94 & 1.44 & 2.143 & 1.654 & 0.306 & 0.478 \\
		Al$_2$O$_3$ & 2.85 & 4.77 & 2.66 & 3.279 & 1.138 & 1.913 \\
		FeO$_\text{T}$ & 2.01 & 5.17 & 2.271 & 1.672 & 1.293 & 2.787 \\
		MgO & 1.06 & 4.08 & 2.302 & 1.21 & 0.724 & 0.947 \\
		CaO & 2.65 & 3.07 & 2.317 & 2.283 & 0.805 & 1.244 \\
		Na$_2$O & 0.62 & 2.29 & 0.815 & 1.075 & 0.409 & 0.745 \\
		K$_2$O & 0.72 & 0.98 & 1.32 & 1.841 & 0.536 & 0.796 \\
		\\
		\hline
	\end{tabular}
	}
\end{table*}
Table \ref{tab:table3} shows that the pre-processing RMSE performance of the proposed spectral CNN remains approximately constant across the different physical distances at which samples where queried. The interpretation of such result is that the spectral CNN is capable of learning the distance corrections applied by \cite{Wiens:2013} and is thus equipped with some level of invariance. Such invariant capability appears to hold as long as a sufficient number of training examples per distance is available at the learning stage. Note that from 5-7m distances there seems to be a marginal RMSE loss in performance in the test partition. However, we attribute such loss to the estimation uncertainty from using a small sample size; in this case 310 and 88. We conclude the pre-processing experimentation by mentioning that in our implementations the spectral CNN was able to pre-process individual shots in a feed forward pass at rates $\sim$50 Hz on a CPU. Such rates render this method as computationally efficient with real-time deployment capabilities. 


\subsection{Qualitative and quantitative chemical content (QQCC) of samples.}

Experiments were conducted to determine and validate performance of the learned QQCC estimators proposed here and described in Section \ref{Ssec:chemometric_mappings}. Such performance is measured using the 'Calib' dataset containing corresponding examples of raw, pre-processed and ground truth QQCC available from the library of known reference calibration standards in \cite{Clegg:2017}. Here, comparisons are done only on the QQCC of the 8 major chemical elements. However, our work is not restricted to these and additions or removals of elements (including minors) to suit the end-user needs can be incorporated within our framework by adding the corresponding parallel linear regressor channels and with the appropriate re-training.

The methods compared include ICA \cite{Forni:2013, Clegg:2017}, SM-PLS \cite{Anderson:2017, Clegg:2017} against the proposed compositional and end-to-end learning schemes. The proposed compositional learning scheme consists of the pre-processing network Modules-1,2 subsequently followed by the regression layers of Module-3 in Figure \ref{fig:architecture} each trained independently in a non end-to-end fashion. The pre-processing is carried out either using the method of \cite{Wiens:2013} or the proposed residual CNN illustrated Module-1 Figure \ref{fig:architecture}. In the case of the end-to-end learning strategy, Modules-1-3 in Figure \ref{fig:architecture} are concatenated and trained end-to-end to generate QQCC estimate results in a single feed-forward pass given only a single shot raw "un-cleaned" spectrum signal. 

Performance is measured by means of the RMSE with a maximum normalization of 100$\%$. A table summarizing the performance results of the compared methods is included in Table \ref{tab:table2}.
From this table we make the following observations. The comparison between the pre-processing by \cite{Wiens:2013} and the proposed learned residual based CNN pre-processing subsequently followed by the linear regressor layers (i.e., compositional spectral CNN) in columns 3 and 4, respectively, show that these two achieve relatively the same performance and a performance comparable to that of SM-PLS in the first column. This further validates that the proposed method is an effective tool for pre-processing raw spectral signals without the prior requirements of including dark current acquisitions and estimation, IRF while also being computationally efficient when in feed-forward mode. In the case of the end-to-end learning scheme shown in the last two columns of Table \ref{tab:table2}, we include results in two independent data partitions of the 'calib' dataset. In partition I, the training set is of size  $0.6$ times the total number of independent shots in the 'Calib' dataset regardless of whether or not it comes from the same target from all the available independent shot measurements and $0.4$ conforms the test set. In contrast, partition II splits the training and testing subsets by $0.9$ and $0.1$ times the total number of shots in the 'Calib' dataset, respectively. In such random partition, we ensure that there are no independent shots from the same target in any of the two subsets. The findings summarized in Table \ref{tab:table2} support the notion that learning tasks (1) and (2) end-to-end and without pre-processing supervision offers additional performance advantages over its compositional processing counterpart, mainly from avoiding the propagation of errors induced by learning pre-processing transformations under supervision. 
In terms of computational complexity and network size, the added requirements of concatenating the linear regressor layers to the pre-processing CNN were insignificant. In fact, both compositional and end-to-end learning schemes can generate QQCC estimate results also at rates of $\sim$50 Hz on CPU with memory storage requirements of $\sim$1MB per model. 

One thing to note a side from the achieved performance and with regards to both the SM-PLS and ICA methods listed in the 2nd and 3rd columns of Table \ref{tab:table2}, respectively is that these numbers were obtained directly from \cite{Clegg:2017}. In addition, the inputs both in training and testing for such ICA and SM-PLS methods make use of averaged spectra over multiple shots in a single location (e.g., 45, with first 5 shots disregarded due to the high presence of adsorbed H$_2$O and possible contamination on the samples). In contrast, the proposed method uses individual shot spectrum signals and is thus capable of disentangling the spectral signal from the sources of uncertainty. In the following paragraphs, we detail the process by which each method is performed and compared.

\subsubsection{Learning by composition} \label{SSec:disjoint}
For learning by composition, we refer to the sequential pre-processing under supervision followed by the QQCC estimator also under supervision both trained independently. The two pre-processing methods compared to perform task (1) are: the method of \cite{Wiens:2013} and the learned spectral CNN proposed here and trained as described in Section \ref{SSSec:level1b}. For the QQCC estimation task (2), we used the parallel stack of linear regressor layers in Module-3 of Figure \ref{fig:architecture} on top of the result of each of the compared pre-processing methods. One independent linear regressor module is trained per compared pre-processing method using as input the result of pre-processing either by \cite{Wiens:2013} or in a feed-forward pass through the proposed spectral CNN and as label the corresponding ground truth QQCC of the sample. 

The linear regressor layers are composed of 8 parallel channels each operating to regress a single chemical element. Note that more parallel channels can be added or removed trivially depending on the number of major and/or minor chemical elements being estimated. For training the regressor layers, the 'Calib' dataset was partitioned into randomly drawn $\sim47,000$ and $\sim31,000$ training and testing independent subset examples, respectively.

\subsubsection{Learning end-to-end} \label{SSec:joint}
\begin{figure*} [t]
	\centering 
	\subfloat[$\text{SiO}_2$] { \includegraphics[width=0.23\linewidth]{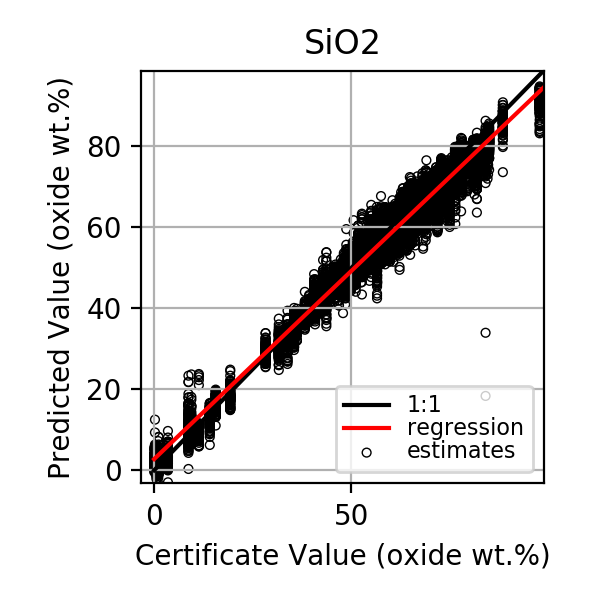} }
	\subfloat[$\text{TiO}_2$]{ \includegraphics[width=0.23\linewidth]{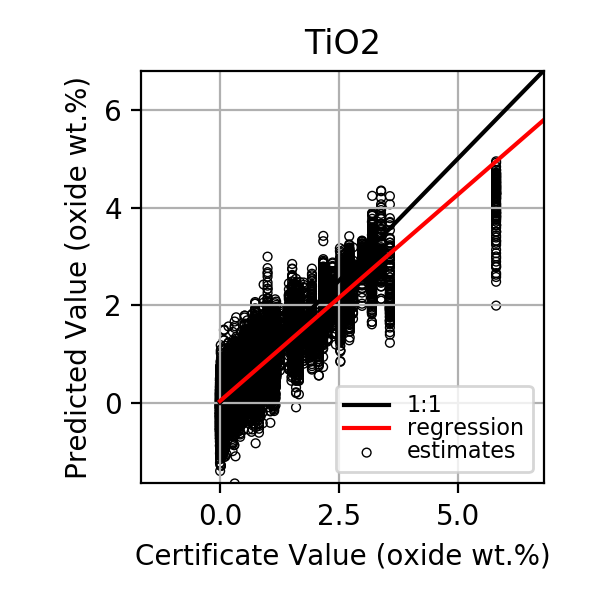} }
	\subfloat[$\text{Al}_2\text{O}_3$]{ \includegraphics[width=0.23\linewidth]{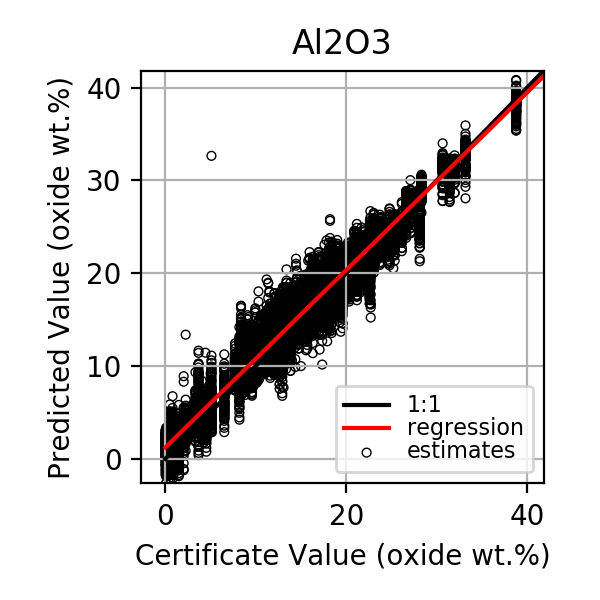} }
	\subfloat[FeO$_\text{T}$]{ \includegraphics[width=0.23\linewidth]{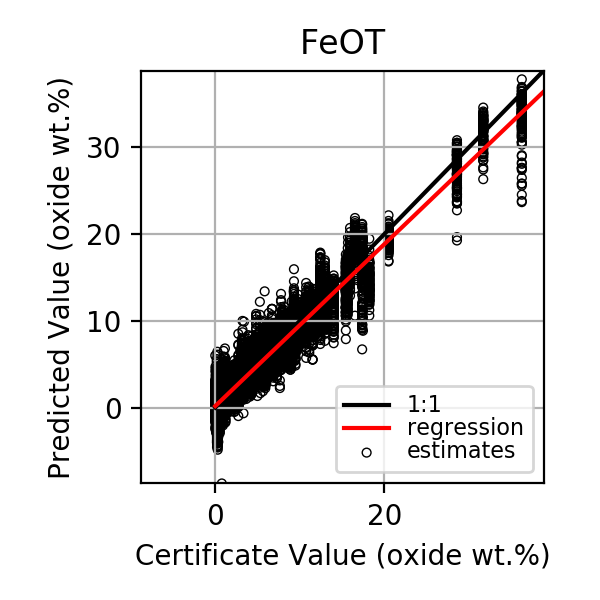} }
	
	\subfloat[MgO]{ \includegraphics[width=0.23\linewidth]{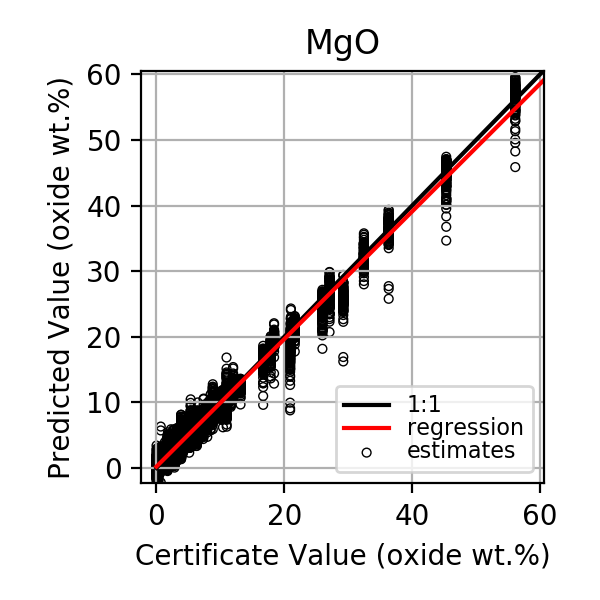} }
	\subfloat[CaO]{ \includegraphics[width=0.23\linewidth]{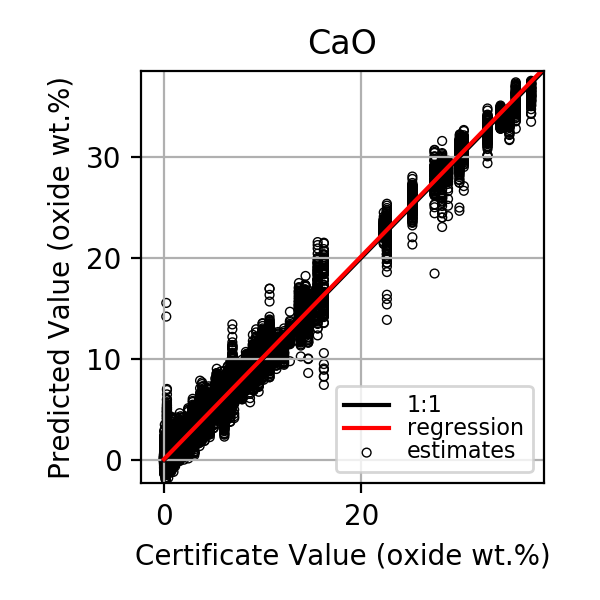} }	
	\subfloat[$\text{Na}_2\text{O}$]{ \includegraphics[width=0.23\linewidth]{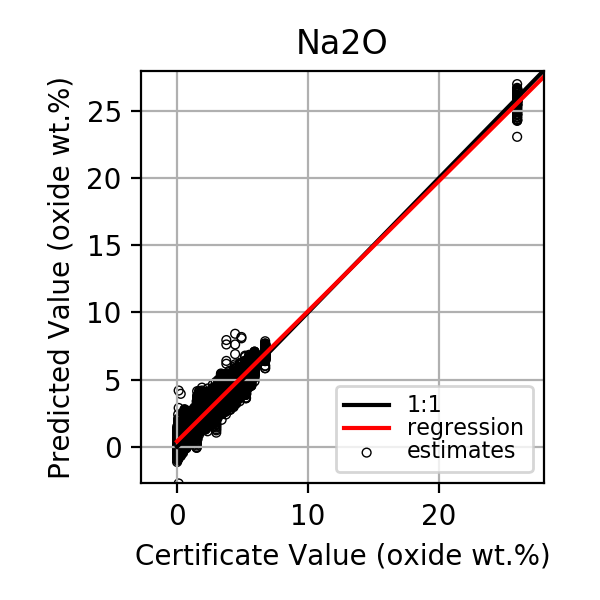} }
	\subfloat[$\text{K}_2\text{O}$]{ \includegraphics[width=0.23\linewidth]{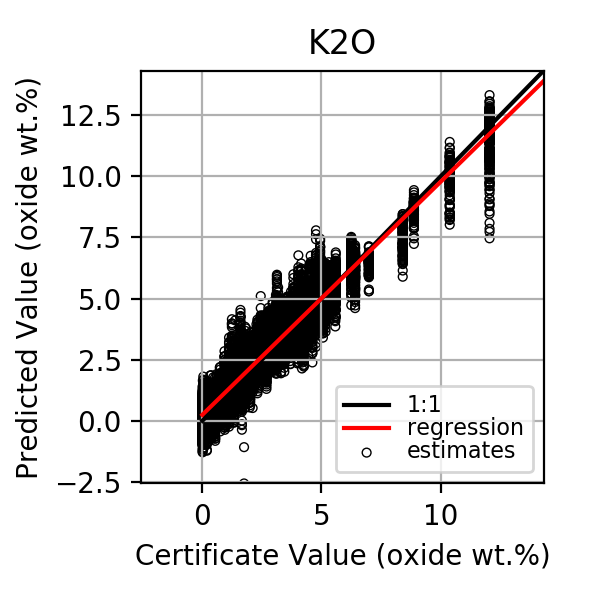} }	
	\caption{Learning qualitative and quantitative chemical content of samples: raw (single-shot) spectral signal to chemical element composition. Regression results of the composition/concentration of each of the 8 major chemical elements in the $\sim31,000$ shot test partition 'Calib' dataset. Corresponding RMSE's for each element are shown in second to last column of Table \ref{tab:table2}.}
	
	\label{fig:concentration} 
\end{figure*}
The experiments conducted to validate the proposed end-to-end learning scheme to estimate QQCC given a (single shot) raw spectral signal consists of measuring RMSE performance for each of the 8 major elements. Training was performed in an end-to-end fashion using pairs of (single shot) raw and corresponding major element concentrations in the 'Calib' train set as input and label, respectively. Partition I used used a training set of $\sim 47,000$ randomly drawn (single-shot) examples while the test set is comprised of $\sim 31,000$ independent examples. Partition II instead splits the dataset into a training set of $\sim 54,000$ randomly drawn (single-shot) examples and a testing set comprised of $\sim 7,500$ drawn example shots ensuring that no shots coming from the same calibration target exists in both subsets.

A summary of the resulting predictions of QQCC for the test set of $\sim 31,000$ single shot raw (edr) inputs in partition I is shown in Figure \ref{fig:concentration}. The horizontal axis corresponds to the ground truth certificate value whereas the vertical axis shows the prediction with the spectral CNN. The red line corresponds to the regression of the point estimates shown in black and unfilled circles while the black 1:1 line is the optimal regression. Note that in general most point estimates follow the 1:1 line in all eight chemical element cases, that the spread of points around the regressed lines is relatively small and that the regressed red lines are in general close to the 1:1 line.  Chemical elements with larger regressed line deviation from the 1:1 line are those with no certificate value in larger oxide wt.$\%$ portions plausibly causing small miss-alignments at those regions. Such is the case for TiO$_2$ for which no high oxide wt. $\%$ is available in the reference calibration standards \cite{Clegg:2017}. However, even in such case the resulting spread as measured by the RMSE summarized in the last column of Table \ref{tab:table2} is significantly small.


\section{Conclusion}
\label{Sec:conclusion} 
This paper proposes a deep spectral convolutional neural network (CNN) to learn to (1) disentangle spectral signal from sources of sensor uncertainty and effects and (2) estimate qualitative and quantitative chemical content (QQCC) of sample given its LIBS signatures. The results of our experimentation show that the proposed deep spectral CNN architecture achieves high precision in learning the standard pre-processing used by the Mars Science Lab while being ready for real-time deployments and offering additional practical benefits. Such practical benefits include eliminating the requirements for dark current measures, instrument response function availability and side environment information such as temperature and emitter-to-target range automatic adjustments when in feed-forward mode. For the task of QQCC estimation, we found that the proposed learning by composition scheme is comparable in performance to existing techniques while offering the additional benefits of imposing less prior information requirements (e.g., dark current acquistions) by learning, requiring only a single laser shot at a location and being ready for real-time deployments. In the case of the proposed end-to-end learning scheme our findings show that in addition to the benefits of the learning by composition scheme also outperforms to the best of our knowledge the existing techniques available in the literature and used by the Mars Science Lab.

\section*{Acknowledgement}
Research presented in this article was supported by the Laboratory Directed Research and Development program of Los Alamos National Laboratory under project number 20200666DI.


\bibliographystyle{elsarticle-num} 

\end{document}